\documentclass{scrartcl}
\usepackage[utf8]{inputenc}
\usepackage{amsmath}
\usepackage[english]{babel}
\usepackage[margin=1in]{geometry}
\usepackage{setspace}
\usepackage{natbib}
\usepackage{subfig}
\usepackage{graphicx}
\usepackage{makecell}
\usepackage{multicol}
\usepackage{booktabs}
\usepackage{caption}
\usepackage[utf8]{inputenc}
\RequirePackage[table]{xcolor}
\usepackage{subfiles}
\usepackage{multirow}

\definecolor{PANDarkGray}{RGB}{153,153,153}

\usepackage{natbib}
\title{What is Sentiment Meant to Mean to Language Models?}
\author{Michael Burnham\footnote{Department of Political Science and Center for Social Data Analytics\\ Pennsylvania State University}\\mlb6496@psu.edu}
\date{May 2024 \\Word Count: 3,265}

\begin{document}

\maketitle
\begin{abstract}
\noindent Sentiment analysis is one of the most widely used techniques in text analysis. Recent advancements with Large Language Models have made it more accurate and accessible than ever, allowing researchers to classify text with only a plain English prompt. However, ``sentiment'' entails a wide variety of concepts depending on the domain and tools used. It has been used to mean emotion, opinions, market movements, or simply a general ``good-bad'' dimension. This raises a question: What exactly are language models doing when prompted to label documents by sentiment? This paper first overviews how sentiment is defined across different contexts, highlighting that it is a confounded measurement construct in that it entails multiple variables, such as emotional valence and opinion, without disentangling them. I then test three language models across two data sets with prompts requesting sentiment, valence, and stance classification. I find that sentiment labels most strongly correlate with valence labels. I further find that classification improves when researchers more precisely specify their dimension of interest rather than using the less well-defined concept of sentiment. I conclude by encouraging researchers to move beyond ``sentiment'' when feasible and use a more precise measurement construct.

\end{abstract}
\textbf{Keywords:} 
\clearpage
\doublespacing

\section{Introduction}
Sentiment analysis is perhaps the most widely used technique in text analysis. With the proliferation of transformer language models and zero-shot classification (i.e. classification without supervised training), many have turned to large language models (LLMs) as accessible and high performance sentiment classifiers\cite[e.g.][]{tornberg2023use, weber2023evaluation, rathje2023gpt}. One challenge with this, however, is that ``sentiment'' entails a wide variety of concepts depending on the domain and tools used. This raises a question: If researchers do not have a consistent definition of sentiment, how do LLMs used for sentiment classification understand it? This brief paper answers this question and spells out the implications for researchers. I first provide a brief overview of sentiment analysis and how it is understood across different domains, tools, and contexts. I then test how generative LLMs understand the concept relative to the more well specified concepts of stance (i.e. opinion) and emotional valence. Finally, I provide recommendations on how researchers should or should not leverage LLMs for sentiment analysis.

Textbooks and literature reviews on sentiment analysis often define sentiment in terms of ``opinions, sentiments, and emotions in text'' \citep{liu2010sentiment}, or ``the computational treatment of opinion, sentiment, and subjectivity in text'' \citep{pang2008opinion}. Indeed, the literature is awash with references to sentiment as both opinion, emotion, and other dimensions of text \citep[e.g.][]{wankhade2022survey, mehboob2020sentiment, hutto2014vader, baccianella2010sentiwordnet}. Those familiar with survey research might characterize these definitions as ``double barreled'' -- they are trying to measure two or more things at once. In the language of statistical inference, sentiment is a ``confounded'' measure -- it encapsulates multiple concepts simultaneously and it is difficult to determine what each concept is contributing to the measure. 

I am not the first to call attention to sentiment's confoundedness. As several other researchers have pointed out, emotions and opinions are two different things that need not align \citep{aldayel2019your, aldayel2021stance, bestvater2022sentiment}. Within computer science, the widely regarded Semeval series of annual workshops has similarly taken care to distinguish between emotion (which they call sentiment) and opinion (which they call stance) \citep{mohammad2016semeval, rosenthal-etal-2017-semeval}.

More broadly, however, such care is not always attended to when using sentiment as a measurement instrument. The commonly used sentiment dictionary Vader, for example, explicitly aligns itself with emotional analysis as a ``valence aware dictionary'', then describes itself as an ``opinion mining'' tool \citep{hutto2014vader}. SentiWordNet similarly positions itself as a tool for ``sentiment classification and opinion mining'' \citep{baccianella2010sentiwordnet}. Other dictionaries, like the popular Linguistic Inquiry and Word Count, avoid any mention of opinion classification and focus entirely on emotional dimensions \citep{pennebaker2001linguistic}. Others still define sentiment in domain specific terms such as ``consumer sentiment'' or what direction the economy is expected to move \citep{yang2020finbert, sousa2019bert}. Finally, several tools evade the question altogether and define sentiment in terms of an undefined positive and negative dimension \cite[e.g.][]{rinker2017package, young2012affective}. This approach reduces sentiment to an ambiguous factor that that correlates with numerous dimensions of interest, yet lacks the precision ideal for scientific inquiry.

The confounded nature of sentiment has resulted in the widespread use of sentiment analysis in cases where it is unclear how valid the measurement approach is. Researchers have used emotion identification to measure opinion inversion \citep{matalon2021using}, applied sentiment analysis without explicitly defining what they are measuring \citep{osmundsen2021partisan}, and in once case, used a model trained to classify product review sentiment on a one to five star scale for analyzing civility on social media \citep{avalle2024persistent}.

Due to recent advancements in LLMs, many have proposed generative AI as a way for robust sentiment classification without training a supervised classifier \citep{tornberg2023use, weber2023evaluation, rathje2023gpt}. These models are attractive tools because they provide state-of-the-art performance without training a supervised classifier, and are highly accessible because of their low technical barrier to entry. The models work by users providing a plain text description of the text analysis task, and then the model executes the task based on the description. However, the ambiguous nature of sentiment outlined above raises an important question: How do large language models understand the task of ``sentiment analysis'' when prompted to do it? If the model's understanding of sentiment is not consistent with the researcher's, they may not be measuring what they think they are. Further, if sentiment is a confounded measurement, would text analysis improve if we disambiguate the term and measure more precise concepts? In the following sections I present answers to these questions.

\section{Data and Methods}
I approach these questions by using LLMs to classify documents three different times: once for sentiment, once for emotional valence, and a final time for stance (opinion). I then evaluate which of the classification approaches is best able to recover manual labels from two data sets, and examine how well sentiment classifications correlate with both opinion and emotional valence.

The first data set I use consists of 2,390 hand labeled tweets about politicians. The labels indicate whether the text expresses support, opposition, or no opinion towards the target politician. The sentiment dataset is a random sample of 2,000 tweets from the test set for task 4 of the 2017 SemEval challenge \citep{rosenthal-etal-2017-semeval}. This data consists of tweets that have been labeled for positive, negative, or neutral sentiment with sentiment being explicitly defined as emotional valence. The data was sampled to have an equal distribution between sentiment classes.

The generative LLMs I test are GPT-4 Turbo, Claude-3 Opus, and Llama-3 8B \citep{openai2023gpt4, claude3report, llama3modelcard}. These three models represent two commonly used proprietary state-of-the-art models in GPT-4 and Opus, as well as a smaller, open-source model that can feasibly run on local hardware. I use a single set of prompts for all three models: One prompt that describes and requests stance classification, a second prompt that describes and requests emotional valence classification, and finally a third prompt that simply asks for sentiment classification without specifying what sentiment is. The sentiment prompt is a slight adaptation of the prompt tested by \citet{rathje2023gpt}.

I use Matthew's Correlation Coefficient (MCC) as the primary performance metric due to its robustness relative to F1 and ROC AUC \citep{chicco2023matthews}. MCC is a special case of the Pearson correlation coefficient and can be interpreted similarly. It ranges from -1 to 1 with 0 indicating no correlation between true class and estimated class.

\section{Results}
\subsection{Stance}
\begin{figure}
    \centering
    \includegraphics[width = \textwidth]{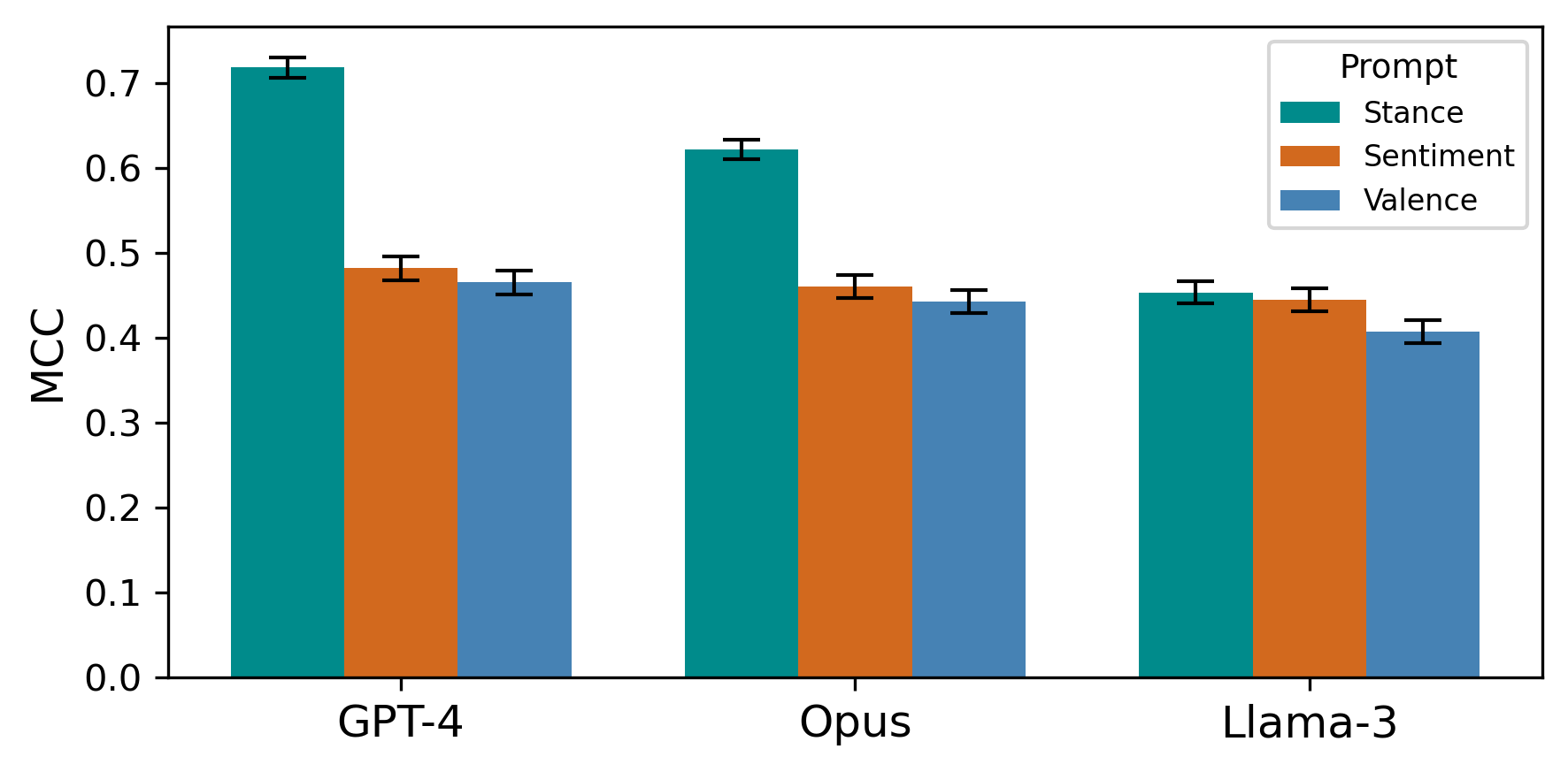}
    \caption{Classification performance with stance, sentiment, and valence prompts on stance labeled documents. Using a prompt for the specific measurement of interest can provide a dramatic improvement over sentiment classification.}
    \label{fig:stance_bars}
\end{figure}

For the stance classification data set I first evaluate how well the stance, sentiment, and valence prompts recover the original labels. Results shown in figure \ref{fig:stance_bars} plot MCC between each classification prompt and the true labels with bootstrapped standard errors. These results demonstrate that directly asking for stance classification far outperforms sentiment and valence prompts when using GPT-4 Turbo and Claude Opus. This result is consistent with previous work from \citet{bestvater2022sentiment}, who emphasize the importance of distinguishing between emotional valence and stance. Across all three models we see a similar pattern of results: Prompting for stance classification yields the best results in recovering stance labels while prompting for valence produces the worst. Consistent with the idea that sentiment is a confounded measure of both stance and valence, the sentiment prompt performs between stance and valence prompts on all three models.

Notably, Llama-3 generally struggles to recover stance labels relative to the larger and more sophisticated GPT-4 and Opus models. While tempting to conclude that stance classification necessitates large models because it requires a more nuanced parsing of a document's semantics, other research has shown this not to be the case \citep{burnham2023stance}. Rather, this should be interpreted as evidence that models have different aptitudes across different tasks and should not be used for classification without some validation first.

\begin{figure}
    \centering
    \includegraphics[width = \textwidth]{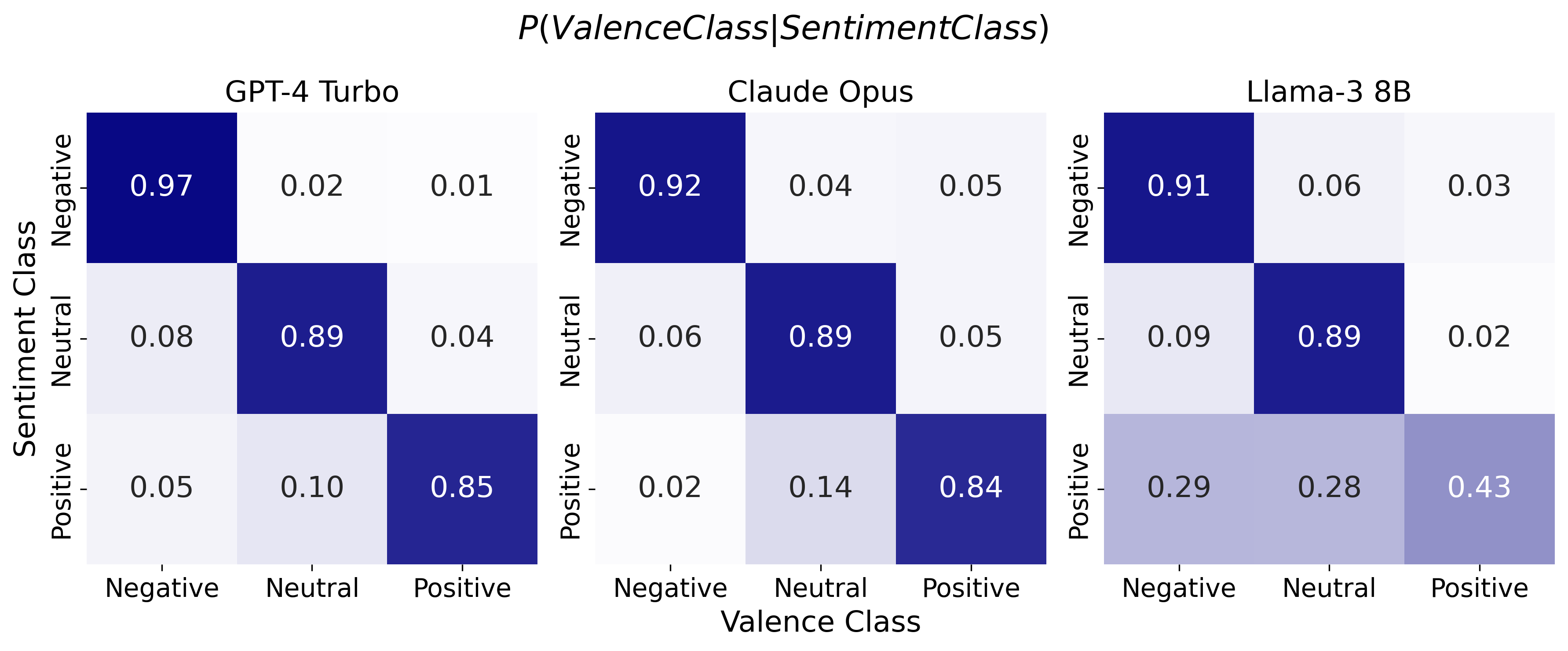}
    \caption{Probability of valence label given sentiment label across models. Higher values indicate greater agreement between valence and sentiment prompts. Valence-sentiment labels agree at a much hugher rate than stance-sentiment labels.}
    \label{fig:sentmat}
\end{figure}

\begin{figure}
    \centering
    \includegraphics[width = \textwidth]{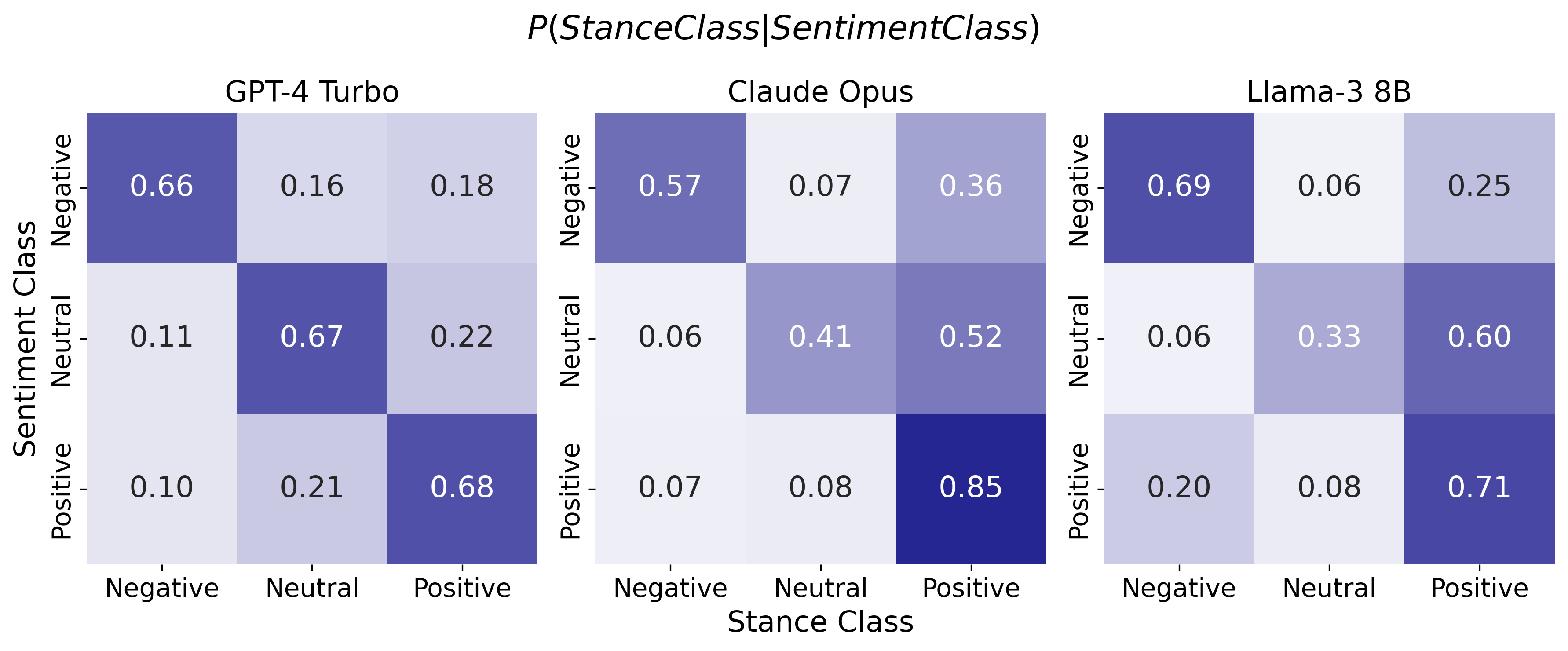}
    \caption{Probability of stance label given sentiment label across models. Higher values indicate greater agreement between stance and sentiment prompts. Stance-sentiment disagreement exceeds valence-sentiment disagreement.}
    \label{fig:stancemat}
\end{figure}

Next, I test how the models understand the concept of ``sentiment'' by comparing the labels produced by the three task prompts. In figures \ref{fig:sentmat} and \ref{fig:stancemat} I present matrices that give the probably of a document's valence or stance class given its sentiment class. The diagonals represent agreement between the two classification approaches, with darker colors correlating with higher probabilities of a valence or stance class given a sentiment class. Solid dark squares on the diagonal and white squares on the off-diagonal would represent perfect agreement between the two classification methods. In figure \ref{fig:sentmat} we see a significant amount of agreement between valence and sentiment across all three models. In each case, the most likely valence class is the same as the sentiment class. Notably, across all three models we see that instances of positive sentiment are the most likely to disagree with the valence class.

Figure \ref{fig:stancemat} demonstrates considerably less agreement between sentiment and stance classes. Claude Opus and Llama-3 in particular show disagreement among documents with neutral sentiment while GPT-4 has considerably lower agreement on the positive class than the other two models. The implication is that if a researcher were to use a sentiment prompt to classify opinion they should expect to recover the correct document label roughly two-thirds of the time at most. This error rate is before even considering miss-classification for other reasons

Based on these results, I find that LLMs generally understand sentiment to mean emotional valence. Researchers looking to conduct opinion mining should use a prompt for opinion classification, not sentiment analysis.

\subsection{Valence}
While LLMs generally understand sentiment in terms of emotional valence, the two are not perfectly correlated. A reasonable hypothesis for this is that LLMs trained on human text have similarly conflated sentiment with various measurement constructs in addition to emotional valence. If this is correct, we would expect the sentiment prompt to perform slightly worse than the valence prompt on data labeled for emotional valence. The more precise valence prompt should help the model better identify the concept we are interested in and alleviate the confoundedness of the sentiment prompt.

\begin{figure}
    \centering
    \includegraphics[width = \textwidth]{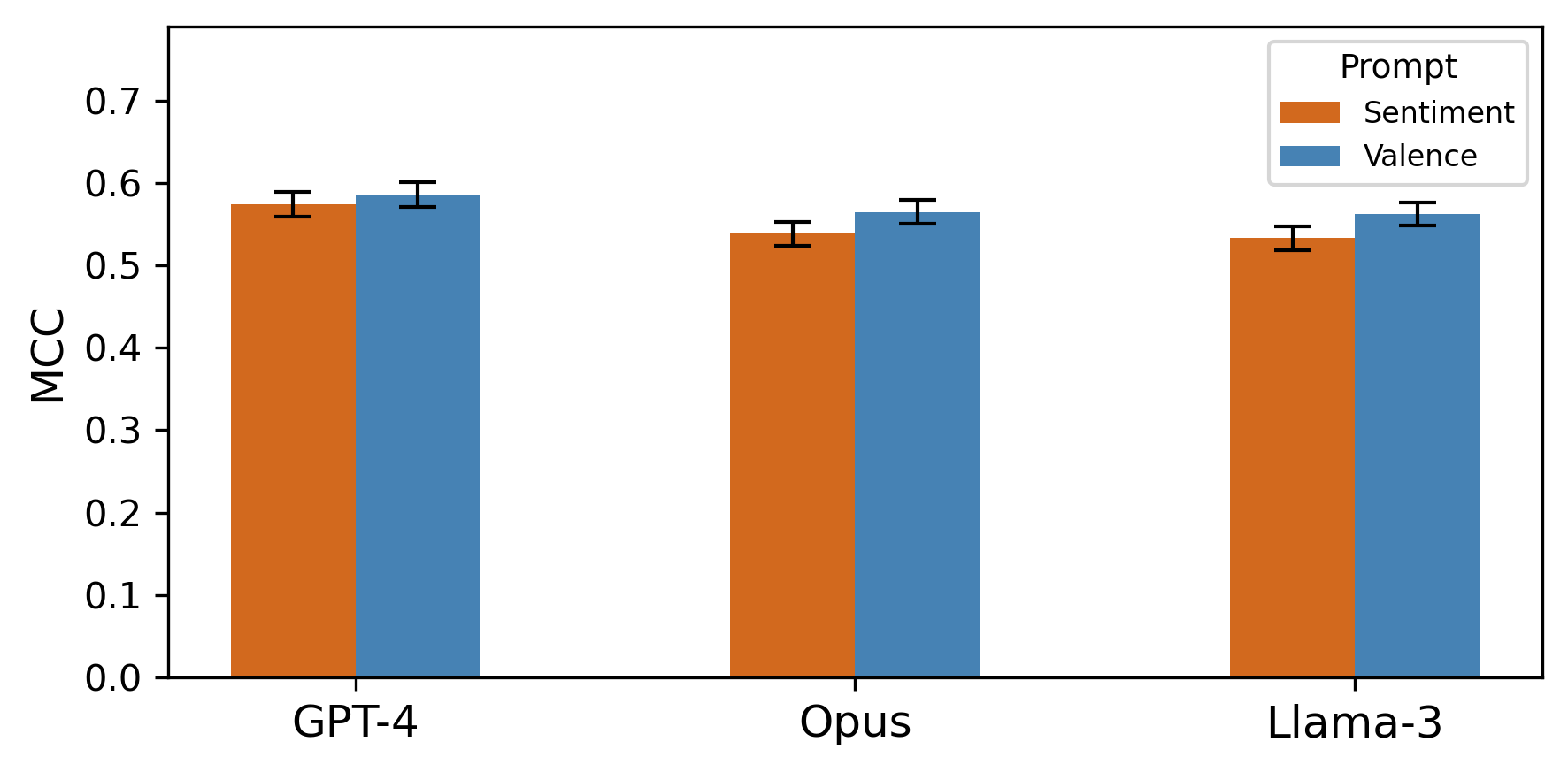}
    \caption{Classification performance on valence labeled documents for sentiment and valence prompts. While LLMs generally understand sentiment sentiment to mean valence, classification performance improves across all models when using a prompt for valence classification.}
    \label{fig:sent_bars}
\end{figure}

To assess this, I classify the SemEval-2017 documents with both the valence and the sentiment prompts. Results in figure \ref{fig:sent_bars} show a small but consistent improvement in performance across models when using the valence prompt rather than the sentiment prompt. This is consistent with the hypothesis that LLMs generally understand sentiment in terms of emotional valence, but that best practice is to be more precise about the dimension of interest rather than relying on the umbrella concept of sentiment.

\section{Discussion}
This paper provided a brief overview of how sentiment is widely used to measure multiple constructs at once, and then demonstrated how LLMs understand the concept. Throughout the literature, researchers define sentiment in terms of emotions, opinions, market movement, a general good-bad dimension, and more. Many of the available tools purport to measure these things simultaneously, and researchers sometimes do not explicitly define sentiment in their work. Given the semantic promiscuity of the term, it remained an open question how generative LLMs understood sentiment analysis and what exactly they do when prompted to classify the sentiment of a document.

The results presented demonstrate that LLMs interpret sentiment primarily as emotional valence, although even valence and sentiment classification did not produce identical results. The implication is that prompting an LLM for sentiment classification can introduce significant measurement error depending on what the researcher is trying to measure. Instead, researchers should direct the model to measure the specific dimension of interest. Even when attempting to measure emotional valence, this paper provides some evidence that researchers should prompt the model to classify valence rather than sentiment. This additional specificity can improve model performance by disentangling the double-barreled nature of ``sentiment.''

Given the above, it is difficult to escape the conclusion that ``sentiment analysis'' is perhaps not the right framing for many text analysis tasks that previously fell under its umbrella. During the first wave of text analysis methods that counted word occurrences with dictionaries, it was difficult to be precise in measurement. Sentiment analysis was a useful approach for measuring many things because emotions can be credibly measured with word counts and correlate with many dimensions of interest. However, the landscape of text analysis has evolved significantly and, fortunately, allows researchers to be much more precise in what they measure. Models like BERT, GPT-4, and others are more capable of modeling document semantics and thus more able to make meaningful distinctions between opinion, emotion, and other dimensions of interest \citep{devlin2018bert, openai2023gpt4, bestvater2022sentiment}. Thus, researchers should strive to use methods that measure these directly rather than bundling them into the single construct of ``sentiment.''

While this paper primarily focused on generative LLMs, it is important to note that such large and computationally demanding models are not necessary to disambiguate between dimensions of sentiment. Encoder models like BERT can also be used as either supervised or zero-shot classifiers to label emotions or opinions with comparable or superior accuracy while requiring much less computing resources \cite{burnham2023stance}.

Accordingly, this paper encourages a shift towards greater precision and clarity in text analysis. When feasible, researchers should explicitly define their dimensions of interest rather the employing the ambiguous umbrella of ``sentiment.'' This will enhance both the validity and reliability of research using text.

\clearpage

\bibliographystyle{chicago}

\bibliography{refs}

\begin{thebibliography}{}

\bibitem[\protect\citeauthoryear{AI@Meta}{AI@Meta}{2024}]{llama3modelcard}
AI@Meta (2024).
\newblock Llama 3 model card.

\bibitem[\protect\citeauthoryear{Aldayel and Magdy}{Aldayel and Magdy}{2019}]{aldayel2019your}
Aldayel, A. and W.~Magdy (2019).
\newblock Your stance is exposed! analysing possible factors for stance detection on social media.
\newblock {\em Proceedings of the ACM on Human-Computer Interaction\/}~{\em 3\/}(CSCW), 1--20.

\bibitem[\protect\citeauthoryear{AlDayel and Magdy}{AlDayel and Magdy}{2021}]{aldayel2021stance}
AlDayel, A. and W.~Magdy (2021).
\newblock Stance detection on social media: State of the art and trends.
\newblock {\em Information Processing \& Management\/}~{\em 58\/}(4), 102597.

\bibitem[\protect\citeauthoryear{Anthropic}{Anthropic}{2024}]{claude3report}
Anthropic (2024).
\newblock The claude 3 model family: Opus, sonnet, haiku.

\bibitem[\protect\citeauthoryear{Avalle, Di~Marco, Etta, Sangiorgio, Alipour, Bonetti, Alvisi, Scala, Baronchelli, Cinelli, et~al.}{Avalle et~al.}{2024}]{avalle2024persistent}
Avalle, M., N.~Di~Marco, G.~Etta, E.~Sangiorgio, S.~Alipour, A.~Bonetti, L.~Alvisi, A.~Scala, A.~Baronchelli, M.~Cinelli, et~al. (2024).
\newblock Persistent interaction patterns across social media platforms and over time.
\newblock {\em Nature\/}~{\em 628\/}(8008), 582--589.

\bibitem[\protect\citeauthoryear{Baccianella, Esuli, Sebastiani, et~al.}{Baccianella et~al.}{2010}]{baccianella2010sentiwordnet}
Baccianella, S., A.~Esuli, F.~Sebastiani, et~al. (2010).
\newblock Sentiwordnet 3.0: An enhanced lexical resource for sentiment analysis and opinion mining.
\newblock In {\em Lrec}, Volume~10, pp.\  2200--2204.

\bibitem[\protect\citeauthoryear{Bestvater and Monroe}{Bestvater and Monroe}{2022}]{bestvater2022sentiment}
Bestvater, S.~E. and B.~L. Monroe (2022).
\newblock Sentiment is not stance: Target-aware opinion classification for political text analysis.
\newblock {\em Political Analysis\/}, 1--22.

\bibitem[\protect\citeauthoryear{Burnham}{Burnham}{2023}]{burnham2023stance}
Burnham, M. (2023).
\newblock Stance detection with supervised, zero-shot, and few-shot applications.
\newblock {\em arXiv preprint arXiv:2305.01723\/}.

\bibitem[\protect\citeauthoryear{Chicco and Jurman}{Chicco and Jurman}{2023}]{chicco2023matthews}
Chicco, D. and G.~Jurman (2023).
\newblock The matthews correlation coefficient (mcc) should replace the roc auc as the standard metric for assessing binary classification.
\newblock {\em BioData Mining\/}~{\em 16\/}(1), 4.

\bibitem[\protect\citeauthoryear{Devlin, Chang, Lee, and Toutanova}{Devlin et~al.}{2018}]{devlin2018bert}
Devlin, J., M.-W. Chang, K.~Lee, and K.~Toutanova (2018).
\newblock Bert: Pre-training of deep bidirectional transformers for language understanding.
\newblock {\em arXiv preprint arXiv:1810.04805\/}.

\bibitem[\protect\citeauthoryear{Hutto and Gilbert}{Hutto and Gilbert}{2014}]{hutto2014vader}
Hutto, C. and E.~Gilbert (2014).
\newblock Vader: A parsimonious rule-based model for sentiment analysis of social media text.
\newblock In {\em Proceedings of the international AAAI conference on web and social media}, Volume~8, pp.\  216--225.

\bibitem[\protect\citeauthoryear{Liu et~al.}{Liu et~al.}{2010}]{liu2010sentiment}
Liu, B. et~al. (2010).
\newblock Sentiment analysis and subjectivity.
\newblock {\em Handbook of natural language processing\/}~{\em 2\/}(2010), 627--666.

\bibitem[\protect\citeauthoryear{Matalon, Magdaci, Almozlino, and Yamin}{Matalon et~al.}{2021}]{matalon2021using}
Matalon, Y., O.~Magdaci, A.~Almozlino, and D.~Yamin (2021).
\newblock Using sentiment analysis to predict opinion inversion in tweets of political communication.
\newblock {\em Scientific reports\/}~{\em 11\/}(1), 7250.

\bibitem[\protect\citeauthoryear{Mehboob, Zaidi, Rizwan, Dilshad, Lashari, Adeel, and Sanwal}{Mehboob et~al.}{2020}]{mehboob2020sentiment}
Mehboob, S., S.~A.~J. Zaidi, M.~Rizwan, U.~Dilshad, N.~Lashari, M.~Adeel, and G.~H. Sanwal (2020).
\newblock Sentiment base emotions classification of celebrity tweets by using r language.
\newblock {\em Pakistan Journal of Engineering and Technology\/}~{\em 3\/}(2), 95--99.

\bibitem[\protect\citeauthoryear{Mohammad, Kiritchenko, Sobhani, Zhu, and Cherry}{Mohammad et~al.}{2016}]{mohammad2016semeval}
Mohammad, S., S.~Kiritchenko, P.~Sobhani, X.~Zhu, and C.~Cherry (2016).
\newblock Semeval-2016 task 6: Detecting stance in tweets.
\newblock In {\em Proceedings of the 10th international workshop on semantic evaluation (SemEval-2016)}, pp.\  31--41.

\bibitem[\protect\citeauthoryear{OpenAI}{OpenAI}{2023}]{openai2023gpt4}
OpenAI (2023).
\newblock Gpt-4 technical report.

\bibitem[\protect\citeauthoryear{Osmundsen, Bor, Vahlstrup, Bechmann, and Petersen}{Osmundsen et~al.}{2021}]{osmundsen2021partisan}
Osmundsen, M., A.~Bor, P.~B. Vahlstrup, A.~Bechmann, and M.~B. Petersen (2021).
\newblock Partisan polarization is the primary psychological motivation behind political fake news sharing on twitter.
\newblock {\em American Political Science Review\/}~{\em 115\/}(3), 999--1015.

\bibitem[\protect\citeauthoryear{Pang, Lee, et~al.}{Pang et~al.}{2008}]{pang2008opinion}
Pang, B., L.~Lee, et~al. (2008).
\newblock Opinion mining and sentiment analysis.
\newblock {\em Foundations and Trends{\textregistered} in information retrieval\/}~{\em 2\/}(1--2), 1--135.

\bibitem[\protect\citeauthoryear{Pennebaker, Francis, and Booth}{Pennebaker et~al.}{2001}]{pennebaker2001linguistic}
Pennebaker, J.~W., M.~E. Francis, and R.~J. Booth (2001).
\newblock Linguistic inquiry and word count: Liwc 2001.
\newblock {\em Mahway: Lawrence Erlbaum Associates\/}~{\em 71\/}(2001), 2001.

\bibitem[\protect\citeauthoryear{Rathje, Mirea, Sucholutsky, Marjieh, Robertson, and Van~Bavel}{Rathje et~al.}{2023}]{rathje2023gpt}
Rathje, S., D.-M. Mirea, I.~Sucholutsky, R.~Marjieh, C.~Robertson, and J.~J. Van~Bavel (2023).
\newblock Gpt is an effective tool for multilingual psychological text analysis.

\bibitem[\protect\citeauthoryear{Rinker}{Rinker}{2017}]{rinker2017package}
Rinker, T. (2017).
\newblock Package ‘sentimentr’.
\newblock {\em Retrieved\/}~{\em 8}, 31.

\bibitem[\protect\citeauthoryear{Rosenthal, Farra, and Nakov}{Rosenthal et~al.}{2017}]{rosenthal-etal-2017-semeval}
Rosenthal, S., N.~Farra, and P.~Nakov (2017, August).
\newblock {S}em{E}val-2017 task 4: Sentiment analysis in {T}witter.
\newblock In S.~Bethard, M.~Carpuat, M.~Apidianaki, S.~M. Mohammad, D.~Cer, and D.~Jurgens (Eds.), {\em Proceedings of the 11th International Workshop on Semantic Evaluation ({S}em{E}val-2017)}, Vancouver, Canada, pp.\  502--518. Association for Computational Linguistics.

\bibitem[\protect\citeauthoryear{Sousa, Sakiyama, de~Souza~Rodrigues, Moraes, Fernandes, and Matsubara}{Sousa et~al.}{2019}]{sousa2019bert}
Sousa, M.~G., K.~Sakiyama, L.~de~Souza~Rodrigues, P.~H. Moraes, E.~R. Fernandes, and E.~T. Matsubara (2019).
\newblock Bert for stock market sentiment analysis.
\newblock In {\em 2019 IEEE 31st international conference on tools with artificial intelligence (ICTAI)}, pp.\  1597--1601. IEEE.

\bibitem[\protect\citeauthoryear{T{\"o}rnberg}{T{\"o}rnberg}{2023}]{tornberg2023use}
T{\"o}rnberg, P. (2023).
\newblock How to use llms for text analysis.
\newblock {\em arXiv preprint arXiv:2307.13106\/}.

\bibitem[\protect\citeauthoryear{Wankhade, Rao, and Kulkarni}{Wankhade et~al.}{2022}]{wankhade2022survey}
Wankhade, M., A.~C.~S. Rao, and C.~Kulkarni (2022).
\newblock A survey on sentiment analysis methods, applications, and challenges.
\newblock {\em Artificial Intelligence Review\/}~{\em 55\/}(7), 5731--5780.

\bibitem[\protect\citeauthoryear{Weber and Reichardt}{Weber and Reichardt}{2023}]{weber2023evaluation}
Weber, M. and M.~Reichardt (2023).
\newblock Evaluation is all you need. prompting generative large language models for annotation tasks in the social sciences. a primer using open models.
\newblock {\em arXiv preprint arXiv:2401.00284\/}.

\bibitem[\protect\citeauthoryear{Yang, Uy, and Huang}{Yang et~al.}{2020}]{yang2020finbert}
Yang, Y., M.~C.~S. Uy, and A.~Huang (2020).
\newblock Finbert: A pretrained language model for financial communications.
\newblock {\em arXiv preprint arXiv:2006.08097\/}.

\bibitem[\protect\citeauthoryear{Young and Soroka}{Young and Soroka}{2012}]{young2012affective}
Young, L. and S.~Soroka (2012).
\newblock Affective news: The automated coding of sentiment in political texts.
\newblock {\em Political communication\/}~{\em 29\/}(2), 205--231.

\end{thebibliography}
%\printbibliography

%\clearpage
%\appendix
%\subfile{Sections/appendix}

\end{document}